\documentclass{article} 

\usepackage{amsmath,amsfonts,bm}

\def\eqref#1{equation~\ref{#1}}

\def\1{\bm{1}}

\DeclareMathAlphabet{\mathsfit}{\encodingdefault}{\sfdefault}{m}{sl}
\SetMathAlphabet{\mathsfit}{bold}{\encodingdefault}{\sfdefault}{bx}{n}

\usepackage{natbib}
\usepackage{hyperref}
\usepackage{url}
\usepackage{amsfonts,amsmath,amssymb, amsthm}
\usepackage{graphicx}
\usepackage{multirow}
\usepackage{booktabs} 
\usepackage{dsfont}
\usepackage{geometry}
\newtheorem{theorem}{Theorem}
\newtheorem{remark}{Remark}
\newtheorem{assumption}{Assumption}
\title{On the Anomalous Generalization of GANs}

\author{Jinchen Xuan \thanks{Equal contribution.} \\
  Peking University \\
  \texttt{1600012865@pku.edu.cn} \\
  Yunchang Yang \footnotemark[1]\\
  Peking University \\
  \footnotetext{Equal distribution.}
  \texttt{1500010650@pku.edu.cn} \\
  Ze Yang \\
  Peking University \\
  \texttt{yangze@pku.edu.cn} \\
  Di He \\
    Key Laboratory of Machine Perception\\MOE\\School of EECS\\Peking University  \\
  \texttt{di\_he@pku.edu.cn} \\
  Liwei Wang \\
    Key Laboratory of Machine Perception\\MOE\\School of EECS\\Peking University  \\
  \texttt{wanglw@cis.pku.edu.cn}
}

\begin{document}
\maketitle
\begin{abstract}
Generative models, especially Generative Adversarial Networks (GANs), have received significant attention recently. However, it has been observed that in terms of some attributes, \emph{e.g.} the number of simple geometric primitives in an image, GANs are not able to learn the target distribution in practice. Motivated by this observation, we discover two specific problems of GANs leading to anomalous generalization behaviour, which we refer to as the sample insufficiency and the pixel-wise combination. For the first problem of sample insufficiency, we show theoretically and empirically that the batchsize of the training samples in practice may be insufficient for the discriminator to learn an accurate discrimination function. It could result in unstable training dynamics for the generator, leading to anomalous generalization. For the second problem of pixel-wise combination, we find that besides recognizing the positive training samples as real, under certain circumstances, the discriminator could be fooled to recognize the pixel-wise combinations (\emph{e.g.} pixel-wise average) of the positive training samples as real. However, those combinations could be visually different from the real samples in the target distribution. With the fooled discriminator as reference, the generator would obtain biased supervision further, leading to the anomalous generalization behaviour. Additionally, in this paper, we propose methods to mitigate the anomalous generalization of GANs. Extensive experiments on benchmark show our proposed methods improve the FID score up to 30\% on natural image dataset. 
\end{abstract}

\section{Introduction}
Generative Adversarial Networks (GANs) have great potential in modeling complex data distributions and have attracted significant attention recently. A great number of techniques \citep{goodfellow2014generative,miyato2018spectral,arjovsky2017wasserstein,gulrajani2017improved,salimans2016improved,brock2018large} and architectures \citep{radford2015unsupervised,karras2018style,zhang2018self,mirza2014conditional} have been developed to make the training of GANs more stable and to generate high fidelity, diverse images. The corresponding generated samples are authentic and difficult for human to distinguish from the real ones.

\begin{figure}[t]
\centering
\includegraphics[width=1.0\linewidth]{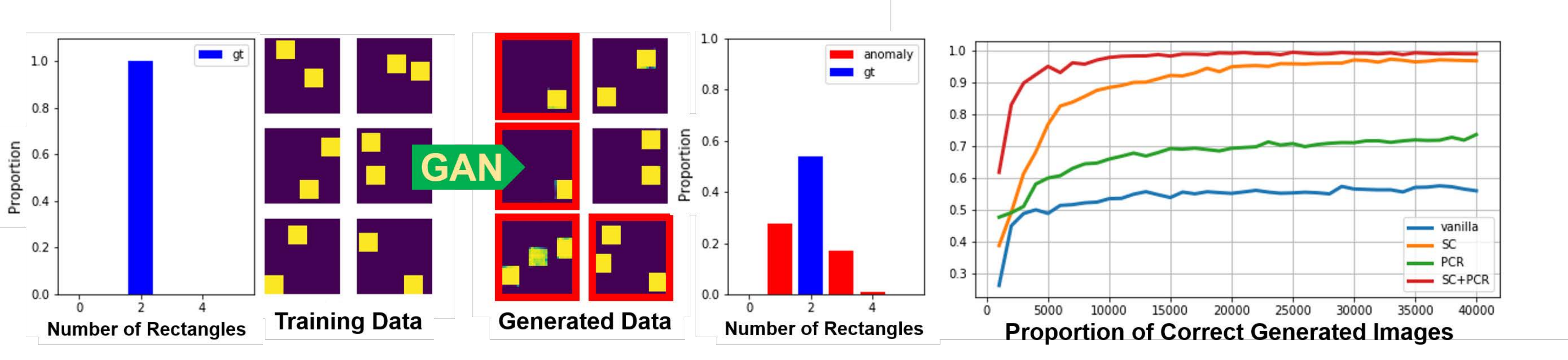}
\caption{Left: The generated images have different rectangle numbers (\emph{e.g.} one, two or three), while the rectangle numbers of all the training data are exactly two (the anomalous marked red). Right: The proportion of the correct generated images (rectangle number is two) for different training approaches. The training dataset consists of 25600 images, all of which have exactly two rectangles.}
\label{fig1}
\end{figure}

Despite these improvements, recent work \citep{zhao2018bias} reported a surprising phenomenon of anomalous generalization of GANs on a geometry dataset, raising new questions about the generalization behaviour. By anomalous generalization it means that several seemingly easy attributes are shown to be learned poorly by GANs, including numerosity (number of objects) and color proportion, which are important for human perception. For example, as shown in Figure \ref{fig1}, for a geometric-object training dataset where the number of objects for each training image is fixed (\emph{e.g.} every training image has exactly two rectangles), it is observed that most generated images after training have very different numbers of objects than the training images (\emph{e.g.} rectangle numbers of most generated images are not two). Mathematically speaking, with regard to the number of objects, the learned distribution of GANs differs significantly from the target distribution, which fails to achieve the goal of modeling the target data distribution faithfully.

Several works have developed theories for GANs. The original work proves the convergence to
equilibrium under ideal conditions \citep{goodfellow2014generative}. Further extensions include \citep{arjovsky2017wasserstein,miyato2018spectral,nagarajan2017gradient,mescheder2018training,bai2018approximability,heusel2017gans}.  \citet{arora2017generalization} points out that GANs may not have good generalization when the discriminator has finite capacity, e.g., neural networks. But they show generalization occurs for GANs under the weak metric of neural net distance. Although these theories provide deep understandings, the generalization and the convergence of GANs as well as how to achieve it in practice are still open problems.


Motivated by this observation, we discover and investigate two specific problems of GANs, namely sample insufficiency and pixel-wise combination, which cause GANs to have anomalous generalization behaviour. Moreover, we propose methods to improve the generalization of GANs.

For the problem of sample insufficiency, we show theoretically and empirically that the batchsize in practice could be insufficient for GANs to model the target data distribution. As a typical setting of GANs, the discriminator learns to separate the fake data distribution of the generator from the real data distribution approximated by the training dataset. In practice, the discriminator learns such a separation function based on the mini-batches sampled from the training dataset and the generated samples of the generator. However, since the size of the mini-batch is much smaller than all the possible samples in the high-dimensional data distribution, the separation function of the discriminator learned based on the mini-batch samples could be noisy. With this noisy discriminator as reference, the generator would learn noisy generation function too and the training dynamics become unstable. As a result, GANs would have anomalous generalization behaviour.    

For the problem of pixel-wise combination, in some situations, we find that the positive training samples and their pixel-wise combinations (pixel-wise average or pixel-wise logical-and) are both recognized as real by the discriminator during training. However, the pixel-wise combinations of the positive training samples could have very different properties from the real samples themselves, indicating that the discriminator is unable to differentiate those seemingly easy attributes (\emph{e.g.} number of objects). With this fooled discriminator as reference, the generator could be fooled further to generate those pixel-wise combinations of training samples, which may not belong to the target distribution. As a result, the data distribution learned by the generator could be very different from the target data distribution and the generalization of GANs becomes anomalous. 

To summarize, our contributions are:
\begin{itemize}
    \item We show that in certain circumstances the discriminator tends to recognize the pixel-wise combinations of the positive training samples as real, which could fool the generator to have anomalous generalization behaviour.
    \item We demonstrate theoretically and empirically that the sample insufficiency in practice could result in unstable training dynamics and anomalous generalization of GANs.
    \item We show that the anomalous generalization reported in \cite{zhao2018bias} is caused by the two problems (sample insufficiency and pixel-wise combination). We then propose novel methods to mitigate anomalous generalization behaviour. Figure \ref{fig1} shows that our proposed methods improve the proportion of correct generated images by almost 80\%. Our methods also improve the FID up to 30\% on natural image datasets.
\end{itemize}
\section{Background}
\subsection{Generative adversarial networks}
In most cases of GANs, the generator learns to map a prior distribution (\emph{e.g.} standard Gaussian) to a fake distribution to approximate the target real data distribution. The discriminator learns a function to separate the real and fake distributions. They define the following min-max game:
\begin{equation}
    \mathop{\text{min}}\limits_{G}\mathop{\text{max}}\limits_{D} \mathop{\mathbb{E}}\limits_{x\sim \mathbb{P}_r}[f_1(D(x))] + \mathop{\mathbb{E}}\limits_{x\sim \mathbb{P}_z}[f_2(D(G(x)))]
\end{equation}
where $\mathbb{P}_r$ and $\mathbb{P}_z$ denote the real and prior distributions respectively. $f_1$ and $f_2$ are the critic functions for the positive and negative training samples (\emph{e.g.} $f_1$(x) = log(x), $f_2$(x) = log(1-x)).

\subsection{Anomalous generalization behaviour of GANs}
Some anomalous generalization behaviours of GANs have been observed recently. In \citet{zhao2018bias}, several seemingly easy attributes are shown to be learned poorly by GANs, including numerosity (number of objects) and color proportion, which are important for human vision systems. The phenomenon shows that the learned distribution of the generator fails to approximate the target distribution, raising new questions about the training dynamics and generalization behaviour of GANs.

\section{Sample insufficiency}
\label{si}
In this section, we first discuss the problem of sample insufficiency in general in Section \ref{si1}. Its empirical observation is shown in \ref{si2}, followed by the theoretical analysis in Section \ref{si3}.

\subsection{Sample insufficiency in the general training of GANs}
\label{si1}
 In the training of GANs, the generator learns to fake the target distribution. The discriminator learns to separate the fake distribution of the generator from the real data distribution. To learn a good separation function between the fake and the real distributions, the discriminator needs to have sufficient information of them. But in practice, such information of the distribution is provided by the positive or negative training samples in the mini-batch. Since the batchsize is often much smaller than the size of all possible data in the high-dimensional data distribution, the information is insufficient and the separation function of the discriminator learned based on the mini-batch samples is noisy. With this noisy discriminator as reference, the generator could also learn a noisy generation function and the training dynamics of GANs become unstable. The smaller the batchsize is, the more unstable the training dynamics are. Since the training of the generator is unstable and the learned generation function is noisy, it is difficult for the learned distribution to approximate the target distribution. As a result, the generalization of GANs would become anomalous. We will show both empirically and theoretically that the sample insufficiency leads to anomalous generalization behaviour of GANs in the following subsections.
 
\subsection{Empirical verification}
\label{si2}
We do experiments to show that the problem of sample insufficiency could lead GANs to anomalous generalization, both for geometric and natural image generation.

\begin{figure}[t]
    \centering
    \includegraphics[width=1.0\linewidth]{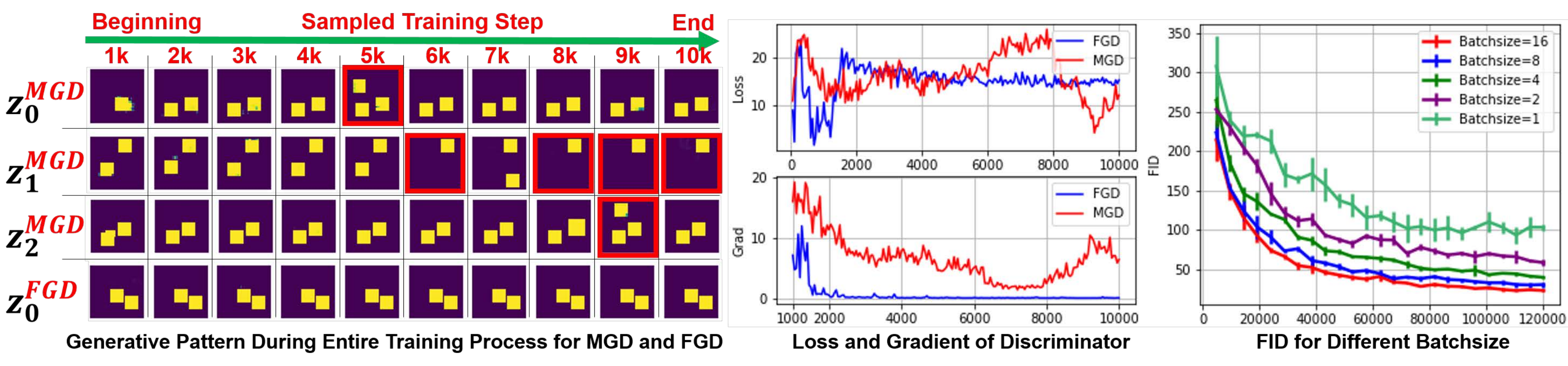}
    \caption{Middle: The loss and gradients for MGD/FGD. Training of MGD is unstable while FGD converges quickly. Left: The generated samples of the generator by four latent codes during training. MGD is unstable and anomalous images are generated ($z_i^{MGD}$, rectangle number is not two). FGD is stable and converges ($z_i^{FGD}$). Right: The FID scores during training for different batchsizes (on CELEB-A). Larger batchsize is better after trained with the same amount of data.}
    \label{asc}
\end{figure}

We establish a geometry dataset consist of 64 images (while the experiment also applies to larger size) that all images (32 by 32 with 0/255 binary pixel value) in it have exactly two rectangles (8 by 8). The prior distribution is the discrete uniform distribution. Size of its support set is the same as the dataset. More details can be found in Appendix \ref{ap_si}. We compare the mini-batch gradient descent (MGD) with the full-batch gradient descent (FGD). The batchsize for MGD and FGD is 16 and 64 respectively.

As shown in Figure \ref{asc} (middle), caused by the sample insufficiency, the training dynamics for the mini-batch gradient descent (MGD) are highly unstable. Both the gradient and the loss go up and down frequently, suggesting it is difficult for GANs to model the target distribution if trained with small batchsize. The instability is also observed for the images generated from certain latent codes, which are the inputs of the generator. As shown in Figure \ref{asc} (left), the generated samples for the three randomly drawn latent codes of MGD ($z_0^{MGD}$,$z_1^{MGD}$,$z_2^{MGD}$) change frequently during training with anomalous images of false rectangle numbers. But for the full-batch gradient descent (FGD), where the sample insufficiency is avoided and the separation function of the discriminator between the real and fake data distributions can be learned accurately at each step, the loss and gradients are stable. The learned distribution converges smoothly to the target distribution in a short time. 

The problem of sample insufficiency is also observed to influence the natural image generation for datasets like CELEB-A \citep{liu2015faceattributes}. As shown in Figure \ref{asc} (right), after trained with the same amount of data, the FID score is better for the larger batchsize, where the problem of sample insufficiency is relatively less severe, than the smaller batchsize, where the sample insufficiency is more problematic. In brief, the experiments show that the sample insufficiency makes the training dynamics unstable, both for the discriminator and the generator. Since the training of the generator and the discriminator are unstable and the learned generation function is noisy, it is hard for the learned distribution to approximate the target distribution. As a result, the final generalization behaviour of GANs becomes anomalous. 

\subsection{Theoretical analysis: a WGAN model}
\label{si3}
In this subsection, we introduce a simple yet prototypical model which shows that the insufficient batchsize will result in unstable training dynamics and smaller batch leads to poorer performance than that of larger batch. As a result, the generalization of GANs would become anomalous when the batchsize is small.
    
Assume that the real data distribution is a $d$-dimensional multivariate normal distribution centered at the origin, $N(\mathbf{0},\mathbf{I}_d)$. The latent distribution of the generator is $p_\mathbf{z}\sim N(\mathbf{0},\mathbf{I}_d)$. The generator is defined as $G_\theta(x)=\theta+x$. The discriminator is a linear function $D_w(x)=w^{\top}x$.

We consider the WGAN model \citep{arjovsky2017wasserstein}, whose value function is constructed using the Kantorovich-Rubinstein duality \citep{villani2008optimal} as
\begin{equation}
W\left(\mathbb{P}_{\theta_{REAL}}, \mathbb{P}_{\theta}\right)=\sup _{\|f\|_{L} \leq 1} \mathbb{E}_{x \sim \mathbb{P}_{\theta_{REAL}}}[f(x)]-\mathbb{E}_{x \sim \mathbb{P}_{\theta}}[f(x)]
\end{equation}
where $\|f\|_L$ denotes the Lipschitz constant of the function $f$, namely, $\|f\|_L=\inf\{L:f \text{ is } L\text{-Lipschitz}\}$. And $\mathbb{P}_{\theta_{REAL}}$ denotes the real data distribution, $\mathbb{P}_{\theta}$ denotes the distribution of the generator $G_\theta$, and the supremum is taken over all the linear functions $f_w(x)=w^{\top}x$ since the optimal classifier $f^*$ is absolutely linear. And when $f_w$ is a linear function, $\|f_w\|_{L}=\|w\|$, so
\begin{align}\label{wgan}
W\left(\mathbb{P}_{\theta_{REAL}}, \mathbb{P}_{\theta}\right)&=\sup _{\|w\| \leq 1} \mathbb{E}_{x \sim \mathbb{P}_{\theta_{REAL}}}[f_w(x)]-\mathbb{E}_{x \sim \mathbb{P}_{\theta}}[f_w(x)]\\
&=\sup _{\|w\| \leq 1} w^{\top}(\mathbb{E}_{x \sim N(\mathbf{0},\mathbf{I}_d)}x-\mathbb{E}_{z \sim N(\mathbf{0},\mathbf{I}_d)}[x+\theta])
\end{align}
The generator is trained to minimize $W\left(\mathbb{P}_{\theta_{REAL}}, \mathbb{P}_{\theta}\right)$. Denote $F(w,\theta)=w^{\top}(\mathbb{E}_{x \sim N(\mathbf{0},\mathbf{I}_d)}x-(\mathbb{E}_{y \sim N(\mathbf{0},\mathbf{I}_d)}y+\theta))$. When we use stochastic gradient descent, the training procedure can be described as
\begin{gather}
    w_{t+1}=w_{t}+\eta_t \nabla_w F(w,\theta_t)\\
    \theta_{t+1}=\theta_t-\mu_t \nabla_{\theta}  F(w_{t+1},\theta)
\end{gather}
where $\eta_t$ and $\mu_t$ are the step size. We present our result for gradient flow \citep{du2018algorithmic,du2018gradient}, i.e., gradient descent with infinitesimal time interval, whose behaviour can be described by the following differential equations:
\begin{align}
\left( \begin{array}{c}{\dot{w}(t)} \\ {\dot{\theta}(t)}\end{array}\right)=&\left( \begin{array}{c}\eta_t \nabla_w F(w(t),\theta(t)) \\ -\mu_t\nabla_{\theta} F(w(t),\theta(t))\end{array}\right)
\end{align}

(A detailed explanation of gradient flow and stochastic gradient flow is in Appendix A.) We will show that when the WGAN in (\ref{wgan}) is trained using stochastic gradient flow, the batchsize will impact the behaviour of the training dynamics. That is, compared with large batchsize (or even full batch), when the batchsize is small, the training dynamics of WGAN suffer from a large variance, thus more unstable. Theorem 1 tells us that when the WGAN model is trained using constant step size stochastic gradient flow, then the variance of the output will increase as $t$ grows, and is of order $\Theta(\frac{1}{m})$. Namely, the variance will be large if the batchsize is small.

\begin{theorem}\label{thm}[Variance of WGAN, constant step size]
Denote $[\theta_t]_i$ as the $i$-th component of the vector $\theta_t$. Suppose we train the WGAN model in (\ref{wgan}) using constant step size stochastic gradient flow with batchsize $m$, then the parameter of the generator $\theta_t$ satisfies
\begin{equation}
\operatorname{Var}([\theta_t]_i)=\int_{0}^{t} \|\left[e^{sA} \Sigma\right]_{d+i}\|_2^2 d s=\Theta(\frac{1}{m})
\end{equation}
where 
\begin{equation}
A=\begin{pmatrix}
0 & -I_d\\
I_d & 0
\end{pmatrix},
\Sigma=\begin{pmatrix}
\sqrt{\frac{2}{m}}I_d & 0\\
0 & 0
\end{pmatrix},
e^{sA}=\begin{pmatrix}
(\cos{s})I_d & -(\sin{s})I_d\\
(\sin{s})I_d & (\cos{s})I_d
\end{pmatrix},
\end{equation}
and $\left[e^{sA} \Sigma\right]_{d+i}$ is the $(d+i)$-th row of the matrix $e^{sA} \Sigma$ ,
\end{theorem}

The proof is in Appendix B. The main idea is that due to the special dynamics of optimization of mini-max problems, the bias caused by the randomness of the batch sampling in each epoch will accumulate, which will lead to large variation after many steps of training. Note that although in traditional optimization problems the variance of SGD caused by the randomness of samples will not affect the convergence \citep{bubeck2015convex,brutzkus2017sgd}, the variance will damage the convergence properties in GANs.

Next we consider the vanishing step size case, in which $\eta_t=\mu_t=1/t$ (without loss of generosity assume $t\geq 1$. This step size is commonly used in convex optimization in practical problems \citep{bubeck2015convex}. Theorem 2 shows that the problem still exists in such case, whose proof is in Appendix C.

\begin{theorem}\label{thm}[Variance of WGAN, vanishing step size]
Suppose we train the WGAN model in (\ref{wgan}) using $1/t$ step size stochastic gradient flow with batchsize $m$, then the parameter of the generator $\theta_t$ satisfies
\begin{equation}
\operatorname{Var}([\theta_t]_i)=\Theta(\frac{1}{m})
\end{equation}

\end{theorem}

\begin{remark}
Here we consider only the WGAN because it is usually more stable than typical GANs. We believe that other forms of GANs will have a similar problem caused by the insufficient batchsize.
\end{remark}

\begin{remark}
Although our theoretical analysis only considers the effect of noise in distribution, the same reason also applies to the effect of other irrelevant features especially when the generator cannot learn the real data distribution perfectly.
\end{remark}

\section{Pixel-wise combination}
In this section, We first discuss the problem of pixel-wise combination in Section \ref{pwc1}, followed by the illustration on toy and real datasets in Section \ref{pwc2}. Finally, we provide a theoretical analysis in Section \ref{pwc3}.

\begin{figure}[t]
\centering
\includegraphics[width=1.0\linewidth]{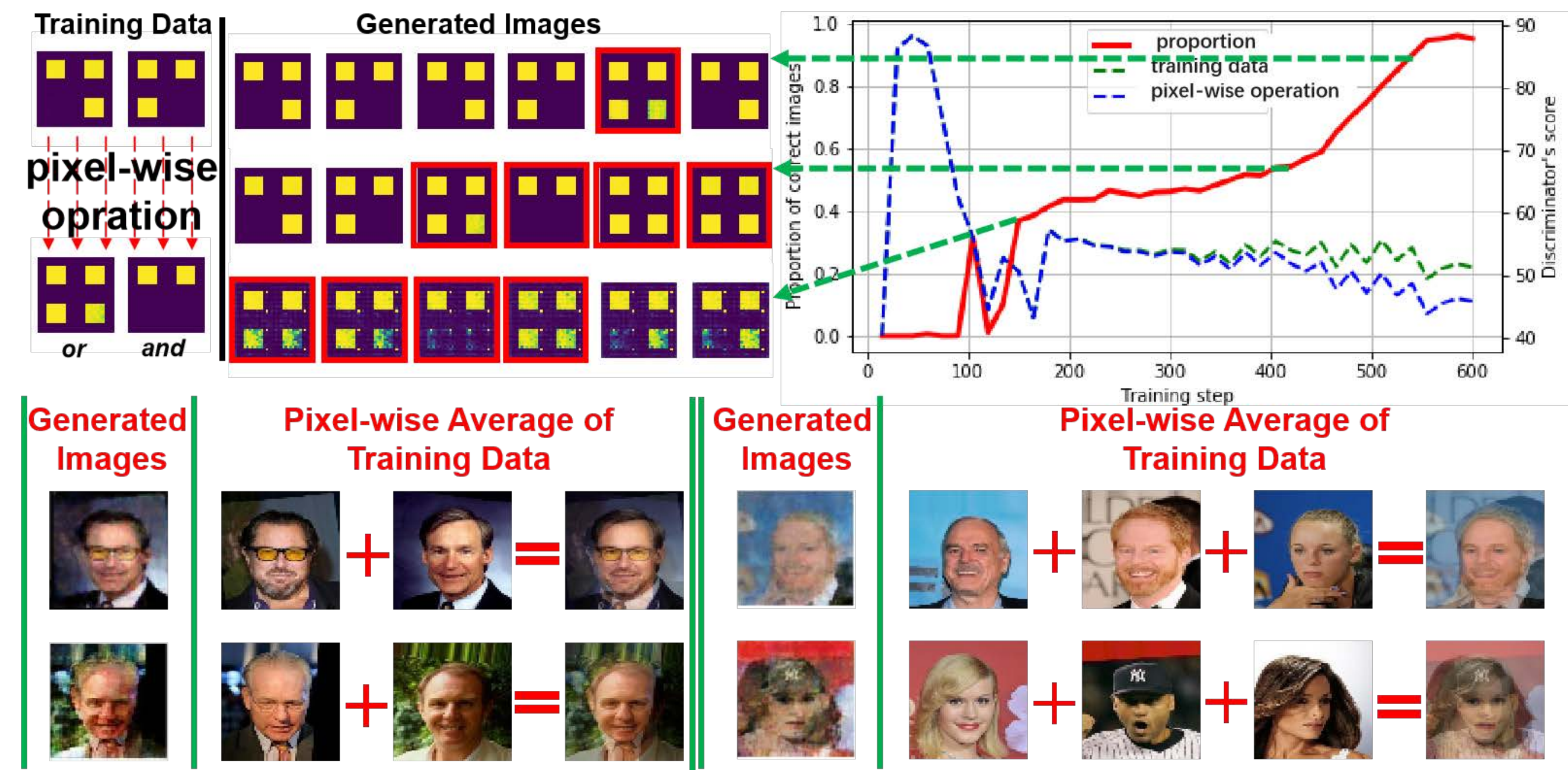}
\caption{Top: For the geometry dataset, the proportion curve of the correct generated images (number of rectangle is three) is plotted (upper right solid line). The scores of the discriminator for the positive training data and the pixel-wise logical-and/or of the positive training data are also plotted respectively (upper right dash line). During training, the generated images which are pixel-wise logical-and/or of the positive training samples prevail (highlighted with colored frame on upper left). They get high scores similar to the positive training images in the earlier stage. Bottom: Certain generated images are shown for human face generation \citep{liu2015faceattributes}. They are exactly the same as the pixel-wise averages of the training data.}
\label{sf}
\end{figure}

\subsection{Pixel-wise combination in the general training of GANs}
\label{pwc1}
When GANs are trained on image datasets, we find that under certain situation pixel-wise combinations (\emph{e.g.} pixel-wise average or pixel-wise logical-and) of the real samples could fool the discriminator, although the generated combinations could have inconsistent properties. \emph{e.g.} the pixel-wise average of two animal images in CIFAR10 could be unrecognizable for human. Take a simple case for illustration, suppose the discriminator is a linear classifier, given two real images, the pixel-wise average of those two real images are likely to be predict as positive sample by the linear discriminator. Since those pixel-wise combinations are recognized as real by the discriminator, the generator correspondingly tends to generate them. Therefore, the generated images could be different from the expected ones and the generalization of GANs becomes anomalous.

\subsection{Illustration on toy and real datasets}
\label{pwc2}
We demonstrate the problem of pixel-wise combination, which leads GANs to anomalous generalization, by a toy dataset. Our training dataset only consists of two binary images (pixel value is 0 or 255), both of which have exactly three rectangles. The positions of the rectangles of the two images are different, as shown at upper left in Figure \ref{sf}. The generated images during training are plotted and their statistics are analyzed. 

As shown at upper left of Figure \ref{sf}, even when the training dataset consists of two images, there are a lot of unexpected anomalous generated samples. Both the two training images have exactly three rectangles. But many generated images have two rectangles or four rectangles. The generated images with three rectangles consist only a small part of all the generated images (Figure \ref{sf} upper right solid red curve). Furthermore, the anomalous generated images are exactly the same as the pixel-wise combinations of the two training images. The anomalous generated images with two rectangles are actually the pixel-wise logical-and of the two training images. The images with four rectangles are actually the pixel-wise logical-or of the two training images. Also, the problem of the pixel-wise combination is observed for the discrimination function of the discriminator. As plotted as dash curves at upper right of Figure \ref{sf}, during the training, the discriminator recognizes the pixel-wise combinations of the two positive training images as real (give high scores by the discrimination function), as well as the two positive training images themselves. With this fooled discriminator as reference, the generator is fooled further to generate unexpected samples. As a result, the learned data distribution could differ a lot from the target data distribution. 

\begin{figure}[t]
    \centering
    \includegraphics[width=1.0\linewidth]{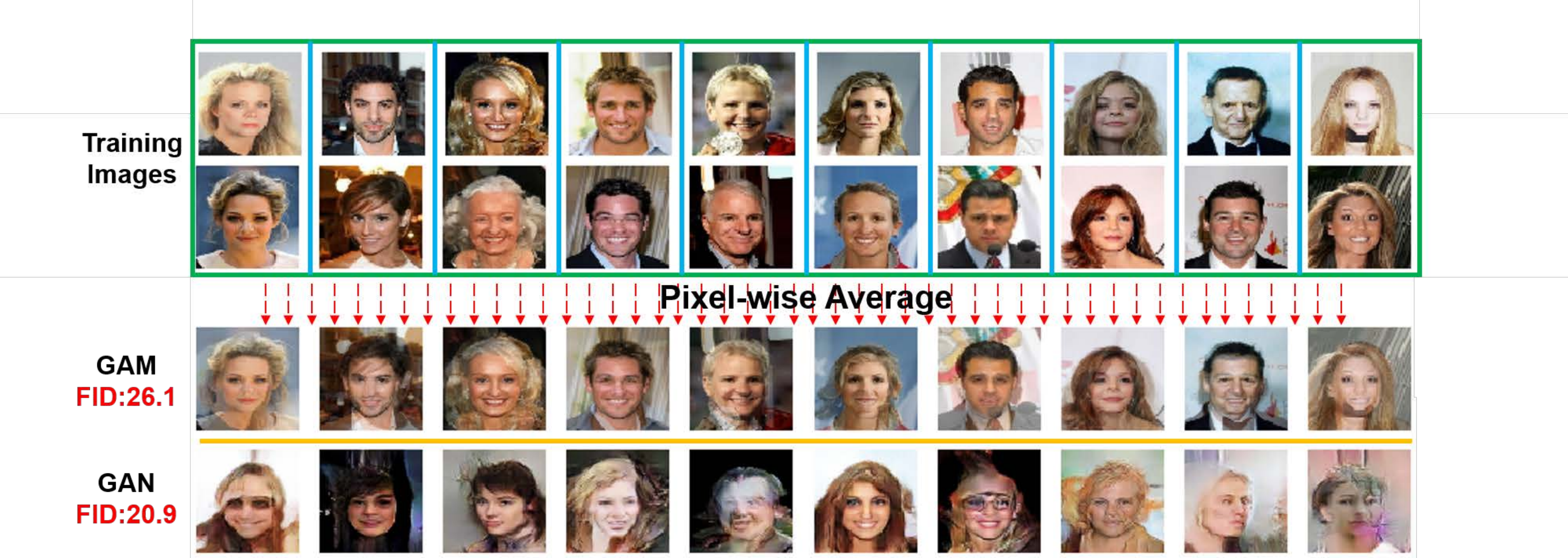}
    \caption{Generated images of the generative average method (GAM) and GANs. The performances are comparable both visually and by the FID score.}
    \label{img_gam}
\end{figure}

Beyond the toy dataset, the problem of pixel-wise combination does exist in the training of GANs in practice. For the natural image dataset CELEB-A, some generated images are exactly the same as the pixel-wise averages of the training data (Figure \ref{sf} bottom). Also, the pixel-wise average of certain structurally similar images could generate realistic samples (Figure \ref{img_gam}). In brief, the experiments show that the problem of pixel-wise combination exists for GANs and makes it hard to model the target data distribution faithfully, which leads to anomalous generalization behaviour.

\subsection{Theoretical analysis}
\label{pwc3}

In this section we give a possible explanation to the problem of pixel-wise combination by a theoretical analysis. Note that this phenomenon is most remarkable on the geometric or the facial datasets. The samples in those datasets are structurally similar, namely, for a facial dataset, the eyes, noses and other features of the human faces appear at fixed positions in the images with high probability. So we assume that the $l_2$ distance between most positive training samples in the dataset is small. 

For the discriminator $f(x)$, without loss of generosity, assume that $f(x) > 0$ if the sample $x$ is classified as real. We make the following assumptions.
\begin{assumption}
Assume that $f$ is $L$-Lipschitz.
\end{assumption} 
\begin{assumption}
Assume that the discriminator classifies all the positive training samples as real with a large margin (\emph{i.e.} there exists $\epsilon>0$ such that for all the positive training samples $x$, $f(x)>\epsilon$).
\end{assumption}
Most classifiers satisfy the Lipschitz condition (for example, the softmax classifier). Assumption 2 means that the discriminator classifies  the positive training samples as real with high confidence. Our theorem shows that under these assumptions, the pixel-wise convex combination of any two positive training samples will be classified as real with high probability. The proof is in Appendix D.

\begin{theorem}
If the discriminator $f(x)$ satisfies Assumption 1-2. Then for any two samples $x_1$ and $x_2$ in the training dataset satisfying $\|x_1-x_2\|_2\leq \delta$ and any $\lambda \in (0,1)$, we have 
\begin{equation}
f(\lambda x_1+(1-\lambda)x_2)\geq \max\{f(x_1)-L(1-\lambda)\delta,f(x_2)-L\lambda\delta\}
\end{equation}
Moreover, if $\epsilon>L\delta\min\{\lambda,1-\lambda\}$, then $f(\lambda x_1+(1-\lambda)x_2)>0$.
\end{theorem}

\section{Fixing the anomalous generalization of GANs}
\label{ftagog}
In this section, we propose novel methods to mitigate the two problems. We present the Pixel-wise Combination Regularization (PCR) to mitigate the problem of pixel-wise combination in Section \ref{pcr}. For the problem of sample insufficiency, we present the Sample Correction (SC) in Section \ref{sc}. The results show that the anomalous generalization for the geometric dataset is avoided entirely (Section \ref{ergd}). For the natural image dataset, the training modifications could improve the FID score up to 30\% (Section \ref{ernd}).
\subsection{Approach}
\label{method}
\paragraph{Pixel-wise Combination Regularization}
\label{pcr}

For the training of vanilla GANs, the positive training samples for the update of the discriminator come from the training dataset and the negative training samples are generated by the generator. Since we think that the discriminator tends to recognize the pixel-wise combinations of the images in the training dataset as real even though they are not in the target distribution, we define a dataset:
\begin{equation}
\label{dcom}
    D_{com} = \{x_0,x_1,...x_{n-1}| x_k = \frac{y_i\oplus y_j}{2}  \ y_i,y_j \in D_{training}\ i\neq j\}
\end{equation}
and use the images in $D_{com}$ as additional negative training samples to restrict this tendency. The $\oplus$ in Eqn. \ref{dcom} is the pixel-wise combination operation, it could be the pixel-wise average or pixel-wise logical-and/or. The samples in $D_{com}$ are the combinations of every two images in the training dataset. The loss term for training with the Pixel-wise Combination Regularization can be written as: 
\begin{equation}
L = \underbrace{\mathbb{E}_{x\sim \mathbb{P}_r}[f_1(D(x))]}_{\text{positive samples}}+ \underbrace{\frac{1}{2}\Big[\mathbb{E}_{x\sim \mathbb{P}_g}[f_2(D(x))] + \mathbb{E}_{x\sim \mathbb{P}_{com}}[f_2(D(x))]\Big]}_{\text{negative samples}}
\end{equation}
where $\mathbb{P}_r$, $\mathbb{P}_g$ and $\mathbb{P}_{com}$ are the data distributions approximated by the training data, the generator and $D_{com}$. In this way, the tendency to generate those combination images is restricted. We refer to this addition of the negative training samples for the training of the discriminator as Pixel-wise Combination Regularization (PCR).

\paragraph{Sample Correction}
\label{sc}
We introduce a general framework to mitigate the problem of sample insufficiency. We assume that to model the target distribution, the discriminator is required to separate accurately the real samples in the target data distribution from the others not in it, which the sample insufficiency makes it difficult to achieve. For that goal, the realistic samples in the negative training batch are not useful. Intuitively, if the realistic samples appear in both the positive and the negative training batches, it would be ambiguous for the discriminator to learn the correct separation function. Therefore, we replace the realistic samples in the negative training batch with less realistic ones by a certain pre-defined measure of reality. In this way, the discriminator could efficiently learn an accurate separation function with limited batchsize. The Sample Correction approach is a general framework and the measure of reality could differ for different datasets. We present our experiments on the geometry and the CELEB-A datasets as examples in the next subsections.
\subsection{Experimental results on geometric dataset}
\label{ergd}
As shown before, caused by the two problems, when trained on a geometry dataset where all the training images have exactly two rectangles, there would be a lot of anomalous generated samples with different number of rectangles. We use the two proposed methods to mitigate this anomalous generalization. We do experiments to verify the effects of our methods. The training dataset consists of 25600 binary images, all of which have exactly two rectangles. For the Pixel-wise Combination Regularization (PCR), the pixel-wise logical-and/or of the positive training images are precomputed (details in Appendix \ref{ap_aag}). They are used as additional negative training samples. For the Sample Correction (SC), the generated realistic samples in the negative training batch of the discriminator are discarded. The realistic samples are those with exactly two rectangles, the same as the positive training samples. 

As shown in Figure \ref{fig1} (right), compared to the vanilla approach, the SC approach (SC) almost eliminates the anomalous generalization and the proportion of correct images (rectangle number is 2) goes up to 97\%. The Pixel-wise Combination Regularization (PCR) approach also improves the proportion but is stuck at 70\%. We think this is caused by the still existence of the problem of sample insufﬁciency. Combining these two methods, the SC+PCR approach converges to 99\%, much more quickly than other approaches, showing the existences of the two problems and the effects of our methods.
\subsection{Experimental results on natural image dataset}
\label{ernd}

We also evaluate the effect of the proposed Pixel-wise Combination Regularization and Sample Correction method on natural image data, where the performances are measured by the FID score. 

For the pixel-wise combination, the pixel-wise averages of the training data are computed simultaneously at each training step of the discriminator. They are used as additional negative training samples. The size of the additional negative training samples is the same as the size of the negative training samples from the generator. The results with the Pixel-wise Combination Regularization (denoted as PCR) are compared with those of the vanilla training based on three popular GANs architectures WGAN-GP, LSGAN and SAGAN \citep{gulrajani2017improved,mao2017least,zhang2018self}. Three natural image datasets are involved: CIFAR10, CELEB-A and M-IMAGENET \citep{liu2015faceattributes,krizhevsky2009learning}. M-IMAGETNET is the validation set of the IMAGENET dataset. We train the network unsupervisedly with the Adam optimizer ($\alpha$ = .0002, $\beta_1$ = .5, $\beta_2$ = .9). As shown in Table \ref{fidtable}, the performances improve in most cases after applying the Pixel-wise Combination Regularization. The FID scores of the baselines are consistent with that reported in \citet{lucic2018gans}. For WGANGP trained on CIFAR10, the achieved best FID score improves up to 30\%, showing the potentials of our regularization method. Interestingly, the improvements are more remarkable for the CIFAR10 and M-IMAGENET than the CELEB-A dataset. We hypothesize this is because the objects in CELEB-A tend to appear at regular or fixed positions. Therefore, the average of real images is likely to give a data point in the target data manifold. For example, it is very possible that the average of two human face images is still a realistic human face image (data in CELEB-A). But it is less possible for the average of images of a car and a dog to be a realistic image (data in CIFAR10).

\begin{table}[!ht]
\caption{The achieved best FID scores of three runs are reported after 50000 steps for different models and datasets. In most cases, the FID score improves after applying the Pixel-wise Combination Regularization (PCR). Positive improvement rates are highlighted in bold. The baseline scores are consistent with that reported in \citet{lucic2018gans}.}
\label{fidtable}
\centering
\begin{tabular}{cccccc}
\toprule
\multicolumn{2}{c}{\multirow{2}{*}{Vanilla/PCR/boost}} & \multicolumn{3}{c}{Model}\\
\cmidrule(lr){3-5}
\multicolumn{2}{c}{~} & WGANGP & LSGAN & SAGAN \\
\midrule
\multirow{3}*{Dataset} & CELEB-A & 20.9 / 21.7 / -3.4\% & 17.7 / 16.0 / \textbf{9.7\%} & 28.0 / 24.3 / \textbf{13.2\%} \\
\cmidrule(lr){2-5}
~ & CIFAR10 & 45.4 / 31.6 / \textbf{30.4}\% & 57.6 / 51.0 / \textbf{11.4\%} & 39.6 / 33.7 / \textbf{14.7\%} \\
\cmidrule(lr){2-5}
~ & M-IMAGENET & 61.8 / 54.1 / \textbf{12.4}\% & 61.0 / 59.4 / \textbf{2.7\%} & 102.9 / 73.4 / \textbf{28.6\%} \\
\bottomrule
\end{tabular}
\end{table}

\begin{figure}[t]
\centering
\includegraphics[width=1.0\linewidth]{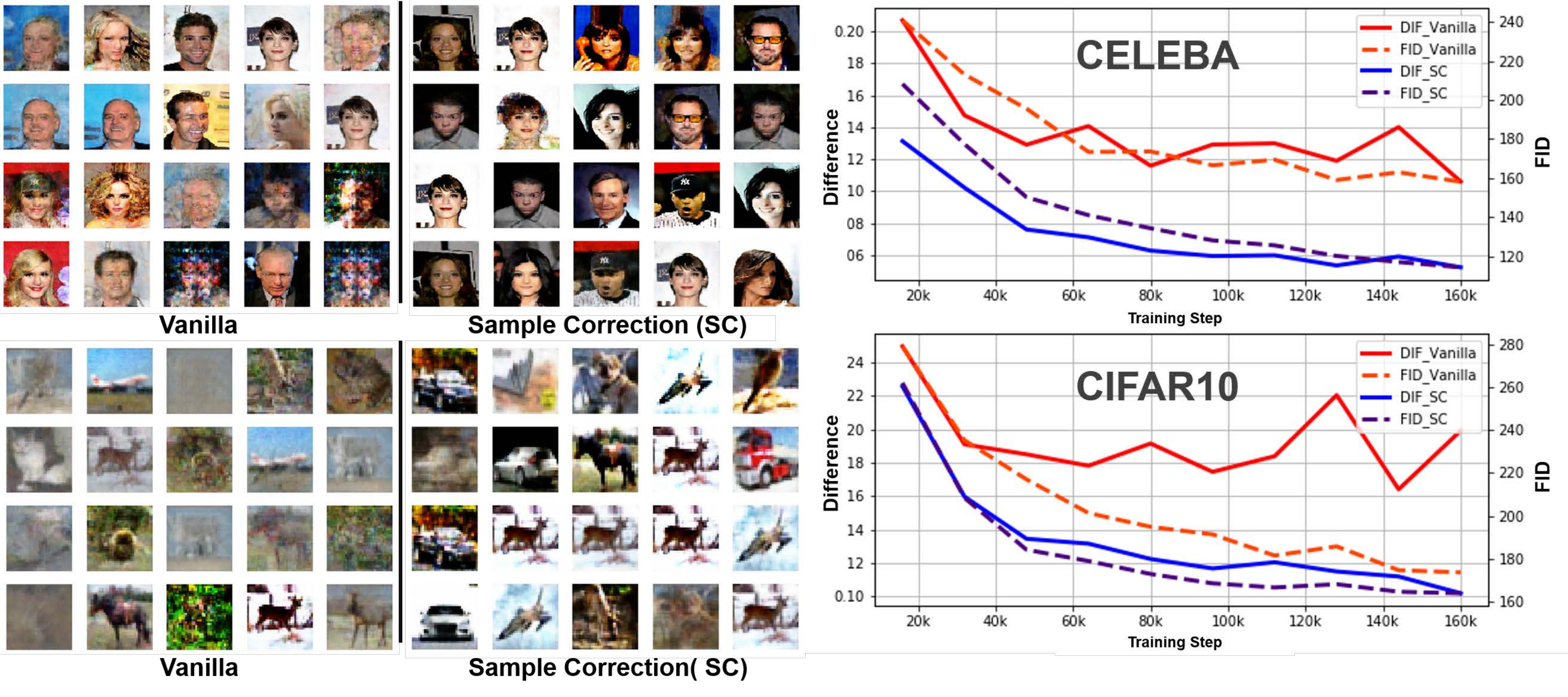}
\caption{Left: The final generated samples for the M-CIFAR10 and M-CELEB-A, where the Sample Correction (SC) approach achieves better performance. Right: The FID and DIF curves during training for the vanilla and the Sample Correction (SC) approaches. }
\label{f2}
\end{figure}

For the Sample Correction method. We randomly select a small portion of the images as training data (M-CELEB-A and M-CIFAR). Dataset size is kept small (\emph{e.g.} 32) so that the learned data distribution could approximate the target data distribution well. The measure of reality of a sample in this case is the normalized minimum $L_2$ distance of the sample to all the training data (DIF, between 0 and 1). We train GANs in two different ways, namely Vanilla and Sample Correction. The latter approach differs in that the realistic samples in the negative batch (DIF less than 0.1) are replaced with less realistic ones. The batchsize is kept small to better demonstrate the problem of sample insufficiency (\emph{e.g.} 2). As shown in Figure \ref{f2}, the Sample Correction approach outperforms the vanilla one by a large margin. FID and DIF scores are both better during training. This is because for the Sample Correction approach, the problem of sample insufficiency is restricted and the separation function of the discriminator is more accurate. The improvements can also be found visually (Figure \ref{f2} left). For the vanilla approach, some generated samples degenerate to noises. But the Sample Correction approach can generate real ones with authentic details.

\section{Conclusion and future work}
In this paper, we discuss two specific problems of GANs, namely sample insufficiency and pixel-wise combination. We demonstrate that they make it difficult to model the target distribution and lead GANs to anomalous generalization. Specific methods are introduced to prevent them from misleading the generalization of GANs, which improve the performance. We hope the two specific problems and the methods to restrict them can help the future work to better understand the generalization behaviour of GANs.

\bibliography{main}
\bibliographystyle{iclr2020_conference}
\clearpage
\appendix

\section{An Explanation of gradient flow and stochastic gradient flow}
Gradient flows, or steepest descent curves, are a very classical topic in evolution equations: given a
functional $F$ defined on $\mathbb{R}^d$, and we want to look for points $x$ minimizing $F$ (which is related to the statical equation $\nabla F(x) = 0)$. To do this, we look, given an initial point $x_0$, for a curve starting at $x_0$ and trying to minimize $F$ as fast as possible. Since the negative gradient direction is the steepest descent direction, we will solve equations of the form 
\begin{equation}
    x'(t) = -\nabla F(x(t))).
\end{equation}
On the curve of the solution of this equation $(t,x(t))$, when $t$ increases, at every point $x(t)$ the point goes along the negative gradient direction.

If we write down the discrete form of (14), it becomes
\begin{equation}
    x(t+\delta t)-x(t) = -\nabla F(x(t)))
\end{equation}
which is the Euler method of the differential equation. And if we take $\delta t=1$, we get the expression of gradient descent. So we can view the gradient flow as gradient descent of infinitesimal time interval.

And if we add a stepsize term in the equation, it becomes
\begin{equation}
    x'(t) = -\eta_t\nabla F(x(t))
\end{equation}
where $\eta_t$ denotes the step size. And in the discrete form, it falls into the familiar gradient descent formula with step size $\{\eta_t\}$:
\begin{equation}
x(t+1)-x(t)=-\eta_t\nabla F(x(t))
\end{equation}

In most machine learning problems, the function $F$ can be written as $F(x)=\frac{1}{n}\sum_{i=1}^n f_i(x)$, and  compute the gradient of $F$ exactly can be computationally exhaustive. So instead we often use an approximation of $\nabla F$ (for example, we can randomly select $i\in \{1,...,n\}$ and use $\nabla f_i$ to approximate $\nabla F$). This can be described formally as follows. If $\nabla F(x(t))=g(x(t))+\mathbb{E}Y$, where $Y$ is a random vector with distribution $p(y)$, then we can sample $Y_t\sim p$ and update $x$ as
\begin{equation}
x(t+1)-x(t)=-\eta_t (g(x(t))+Y_t)
\end{equation}
Especially, if $p$ is a Gaussian distribution with mean 0 and covariance matrix $I_d$, then (18) is equivalent to
\begin{equation}
x(t+1)-x(t)=-\eta_t (g(x(t))+(W(t+1)-W(t)))
\end{equation}
where W(t) is a Wiener process\citep{wiener1923differential}. And this is the Euler-Maruyama scheme\citep{kloeden2013numerical} of the following stochastic differential equation:
\begin{equation}
\begin{cases}
dX_t=-\eta_tg(X_t)dt-\eta_tdW_t\\
X_0=x_0
\end{cases}
\end{equation}
the solution, if exists, is a stochastic process $\{X_t=X(t,x_0)\}$. We call this solution the \emph{stochastic gradient flow} of $F$ at point $x_0$. And when we say that $F$ is trained using stochastic gradient flow, we mean that the curve $\{x_t\}$ is a sample path of $\{X_t=X(t,x_0)\}$. 

Finally, when we say that $F$ is trained using stochastic gradient descent with batchsize $m$, we mean that for each $t$, we sample $Y_t^1,...,Y_t^m \sim p$ and update $x$ as 
\begin{equation}
x(t+1)-x(t)=-\eta_t (g(x(t))+\frac{1}{m}\sum_{i=1}^m Y_t^i).
\end{equation}
This is the Euler-Maruyama scheme of the following stochastic differential equation:
\begin{equation}
\begin{cases}
dX_t^m=-\eta_tg(X^m_t)dt-\eta_t\frac{1}{m}dW_t\\
X^m_0=x_0
\end{cases}
\end{equation}
And when we say that $F$ is trained using \emph{stochastic gradient flow with batchsize $m$}, we mean that the curve $\{x_t\}$ is a sample path of $\{X^m_t=X^m(t,x_0)\}$. 

\section{Proof of Theorem 1}
\label{prf}
\begin{proof}
First we give a detailed description of the training dynamics of our model in Section \ref{si3} using gradient flow. 

The framework of training the WGAN in Section \ref{si3} is as follows. Denote $F(w,\theta)=w^{\top}(\mathbb{E}_{x \sim N(\mathbf{0},\mathbf{I}_d)}x-(\mathbb{E}_{y \sim N(\mathbf{0},\mathbf{I}_d)}y+\theta))$. For each epoch $t=1,2,...$, given the parameter of the generator $\theta_t$, the target of the discriminator is
\begin{equation}
\max_{\|w\| \leq 1} F(w,\theta_t)=w^{\top}(\mathbb{E}_{x \sim N(\mathbf{0},\mathbf{I}_d)}x-(\mathbb{E}_{y \sim N(\mathbf{0},\mathbf{I}_d)}y+\theta_t))
\end{equation}
The gradient of $w$ is 
\begin{equation}
\nabla_w F(w,\theta_t)=\mathbb{E}_{x \sim N(\mathbf{0},\mathbf{I}_d)}x-(\mathbb{E}_{y \sim N(\mathbf{0},\mathbf{I}_d)}y+\theta_t)
\end{equation} 
And for one-step gradient upscent the update of $w$ is
\begin{equation}
    w_{t+1}=w_{t}+\eta_t \nabla_w F(w,\theta_t)
\end{equation}
where $\eta_t$ is the step size.

After $w_t$ is updated to $w_{t+1}$, given the parameter of the discriminator $w_{t+1}$, the target of the generator becomes
\begin{equation}
\min_{\theta} F(w_{t+1},\theta)=w_{t+1}^{\top}(\mathbb{E}_{x \sim N(\mathbf{0},\mathbf{I}_d)}x-(\mathbb{E}_{y \sim N(\mathbf{0},\mathbf{I}_d)}y+\theta))
\end{equation}
And the gradient of $\theta$ is 
\begin{equation}
    \nabla_{\theta} F(w_{t+1},\theta)=-w_{t+1}.
\end{equation}
For one-step gradient descent the update of $\theta$ is 
\begin{equation}
    \theta_{t+1}=\theta_t-\mu_t \nabla_{\theta}  F(w_{t+1},\theta)
\end{equation}

Now we consider the case when $\eta_t$ and $\mu_t$ are constants, $\eta_t=\mu_t=C$. Without loss of generality assume $C=1$. And when the time interval is infinitesimal, the discrete dynamics converge to the gradient flow with constant step size, which can be written in the form of ordinary differential equations (ODEs):
\begin{align}
\left( \begin{array}{c}{\dot{w}(t)} \\ {\dot{\theta}(t)}\end{array}\right)=&\left( \begin{array}{c}\nabla_w F(w(t),\theta(t)) \\ -\nabla_{\theta} F(w(t),\theta(t))\end{array}\right)\\
=&\left( \begin{array}{c}\mathbb{E}_{x \sim N(\mathbf{0},\mathbf{I}_d)}x-(\mathbb{E}_{y \sim N(\mathbf{0},\mathbf{I}_d)}y+\theta(t)) \\ -w(t)\end{array}\right)
\end{align}

Under the full-batch condition, which means that $\mathbb{E}_{x \sim N(\mathbf{0},\mathbf{I}_d)}x=0$ and $\mathbb{E}_{y \sim N(\mathbf{0},\mathbf{I}_d)}y=0$ can be precisely calculated, the solution to the above ODEs is
\begin{equation}\label{ex}
\begin{cases}
w_t=w_0 \cos{t}-\theta_0\sin{t}\\
\theta_t=\theta_0\cos{t}+w_0 \sin{t}
\end{cases}
\end{equation}
Notice that the solution implies that $\|w_t\|\leq\|w_0\|+\|\theta_0\|$, which means that the 1-Lipschitz condition on $w$ in WGAN is automatically satisfied if $w_0$ and $\theta_0$ are initialized sufficiently small. The parameters lie on a circle around the equilibrium point $(0,0)$, which are stable though do not converge.

When the batchsize is $m$, which means that we randomly draw i.i.d. samples $x^1_t,...,x^m_t\sim N(\mathbf{0},\mathbf{I}_d)$ and $y^1_t,...,y^m_t\sim N(\mathbf{0},\mathbf{I}_d)$ and use the sample mean $\bar{x}^{m,t}=\frac{1}{m}\sum_{i=1}^m x^i_t$ and $\bar{y}^{m,t}=\frac{1}{m}\sum_{i=1}^m y_t^i$ to estimate the true mean $\mathbb{E}_{x \sim N(\mathbf{0},\mathbf{I}_d)}x$ and $\mathbb{E}_{y \sim N(\mathbf{0},\mathbf{I}_d)}y$, the gradients become $\nabla_w F=\bar{x}^{m,t}-\bar{y}^{m,t} -\theta$ and $\nabla_{\theta} F=-w$. Since $x^1_t,...,x^m_t$ and $y^1_t,...,y^m_t$ are independent for different $t$, we have $\bar{x}^{m,t},\bar{y}^{m,t}\sim N(0,\frac{1}{m}\mathbf{I}_d)$ and they are independent for different $t$. So $\bar{x}^{m,t}-\bar{y}^{m,t}\sim N(0,\frac{2}{m}\mathbf{I}_d)$. And the parameters $w$ and $\theta$ satisfy the following stochastic differential equations:

\begin{equation}
\begin{cases}
dw_t=-\theta_t dt+\sqrt{\frac{2}{m}}dW_t\\
d\theta_t=w_t dt
\end{cases}
\end{equation}

where $W_t$ is a $d-dimensional$ standard Wiener process. Denote $X_t=(w_t^{\top}, \theta_t^{\top})^{\top}\in \mathbb{R}^{2d}$, then $X_t$ is a multidimensional OU process \citep{uhlenbeck1930theory} with expectation 
\begin{equation}
E\left[X(t) | X(0)\right]=e^{tA}  X(0)
\end{equation}
and variance 
\begin{equation}
\operatorname{Var}\left[X(t) | X(0)\right]=\int_{0}^{t} e^{tA} e^{-sA} \Sigma \Sigma^{\top}\left(e^{tA} e^{-sA}\right)^{\top} d s
\end{equation}
where 
\begin{equation}
A=\begin{pmatrix}
0 & -I_d\\
I_d & 0
\end{pmatrix}
,\Sigma=\begin{pmatrix}
\sqrt{\frac{2}{m}}I_d & 0\\
0 & 0
\end{pmatrix}
\end{equation}
So $e^{tA}=\begin{pmatrix}
(\cos{t})I_d & -(\sin{t})I_d\\
(\sin{t})I_d & (\cos{t})I_d
\end{pmatrix}$ and 
\begin{align}
e^{tA}e^{-sA}=&\begin{pmatrix}
(\cos{t})I_d & -(\sin{t})I_d\\
(\sin{t})I_d & (\cos{t})I_d
\end{pmatrix}\begin{pmatrix}
(\cos{-s})I_d & -(\sin{-s})I_d\\
(\sin{-s})I_d & (\cos{-s})I_d
\end{pmatrix}\\
=&\begin{pmatrix}
(\cos{(t-s)})I_d & -(\sin{(t-s)})I_d\\
(\sin{(t-s)})I_d & (\cos{(t-s)})I_d
\end{pmatrix}=e^{(t-s)A}
\end{align}
So the variance can be written as
\begin{align}
\operatorname{Var}\left[X(t) | X(0)\right]=&\int_{0}^{t} e^{tA} e^{-sA} \Sigma \Sigma^{\top}\left(e^{tA} e^{-sA}\right)^{\top} d s\\
=&\int_{0}^{t} e^{(t-s)A} \Sigma \Sigma^{\top}\left(e^{(t-s)A}\right)^{\top} d s\\
=&\int_{t}^{0} e^{(t-s)A} \Sigma \Sigma^{\top}\left(e^{(t-s)A}\right)^{\top} d (t-s)\\
=&\int_{0}^{t} e^{sA} \Sigma \Sigma^{\top}\left(e^{sA}\right)^{\top} d s
\end{align}
And the variance of the $i$-th component of $X(t)$ is
\begin{align}
\operatorname{Var}\left[X(t) | X(0)\right]_{ii}
=&\int_{0}^{t} \left[e^{sA} \Sigma\right]_i \left[e^{sA} \Sigma\right]_i^{\top} d s\\
=&\int_{0}^{t} \|\left[e^{sA} \Sigma\right]_i\|_2^2 d s
\end{align}
where $\left[e^{sA} \Sigma\right]_i$ denotes the $i$-th row of $e^{sA} \Sigma$. 

Since $\|\left[e^{sA} \Sigma\right]_i\|_2^2\geq 0$, for all $t_1\leq t_2$ we have $\operatorname{Var}\left[X(t_1) | X(0)\right]_{ii}\leq \operatorname{Var}\left[X(t_2) | X(0)\right]_{ii}$. So the variance of the OU process will increase as $t$ grows. And because the elements in $\Sigma$ is of order $\Theta(\frac{1}{\sqrt{m}})$, we have $\operatorname{Var}\left[X(t) | X(0)\right]=\Theta(\frac{1}{m})$. From the definition of $X(t)$ we have $\operatorname{Var}([\theta_t]_i)=\operatorname{Var}[X(t)|X(0)]_{d+i,d+i}=\int_{0}^{t} \|\left[e^{sA} \Sigma\right]_{d+i}\|_2^2 d s$. Hence we complete our proof.
\end{proof}

\section{Proof of Theorem 2}

\begin{proof}
Now we consider the vanishing step size situation, namely $\eta_t=\mu_t=1/t$. And when the time interval is infinitesimal, the discrete dynamics converge to the gradient flow which can be written in the form of ordinary differential equations (assume $t\geq 1$):
\begin{align}
\left( \begin{array}{c}{\dot{w}(t)} \\ {\dot{\theta}(t)}\end{array}\right)=&\left( \begin{array}{c}\frac{1}{t}\nabla_w F(w(t),\theta(t)) \\ -\frac{1}{t}\nabla_{\theta} F(w(t),\theta(t))\end{array}\right)\\
=&\left( \begin{array}{c}\frac{1}{t}(\mathbb{E}_{x \sim N(\mathbf{0},\mathbf{I}_d)}x-(\mathbb{E}_{y \sim N(\mathbf{0},\mathbf{I}_d)}y+\theta(t))) \\ -\frac{1}{t}w(t)\end{array}\right)
\end{align}

Under the full-batch condition, which means that $\mathbb{E}_{x \sim N(\mathbf{0},\mathbf{I}_d)}x=0$ and $\mathbb{E}_{y \sim N(\mathbf{0},\mathbf{I}_d)}y=0$ can be precisely calculated, the solution to the above ODEs is
\begin{equation}
\begin{cases}
w_t=w_1 \cos(\ln t)+\theta_1\sin(\ln t)\\
\theta_t=w_1\sin(\ln t)-\theta_1\cos(\ln t)
\end{cases}
\end{equation}
Notice that the solution implies that $\|w_t\|\leq\|w_1\|+\|\theta_1\|$, which means that the 1-Lipschitz condition on $w$ in WGAN is automatically satisfied if $w_0$ and $\theta_0$ are initialized sufficiently small. The parameters lie on a circle around the equilibrium point $(0,0)$, which are stable though do not converge.

When the batchsize is $m$, which means that we randomly draw i.i.d. samples $x^1_t,...,x^m_t\sim N(\mathbf{0},\mathbf{I}_d)$ and $y^1_t,...,y^m_t\sim N(\mathbf{0},\mathbf{I}_d)$ and use the sample mean $\bar{x}^{m,t}=\frac{1}{m}\sum_{i=1}^m x^i_t$ and $\bar{y}^{m,t}=\frac{1}{m}\sum_{i=1}^m y_t^i$ to estimate the true mean $\mathbb{E}_{x \sim N(\mathbf{0},\mathbf{I}_d)}x$ and $\mathbb{E}_{y \sim N(\mathbf{0},\mathbf{I}_d)}y$, the gradients become $\nabla_w F=\bar{x}^{m,t}-\bar{y}^{m,t} -\theta$ and $\nabla_{\theta} F=-w$. Since $x^1_t,...,x^m_t$ and $y^1_t,...,y^m_t$ are independent for different $t$, we have $\bar{x}^{m,t},\bar{y}^{m,t}\sim N(0,\frac{1}{m}\mathbf{I}_d)$ and they are independent for different $t$. So $\bar{x}^{m,t}-\bar{y}^{m,t}\sim N(0,\frac{2}{m}\mathbf{I}_d)$. And the parameters $w$ and $\theta$ satisfy the following stochastic differential equations:

\begin{equation}
\begin{cases}
dw_t=-\frac{1}{t}\theta_t dt+\frac{1}{t}\sqrt{\frac{2}{m}}dW_t\\
d\theta_t=\frac{1}{t}w_t dt
\end{cases}
\end{equation}

where $W_t$ is a $d$-dimensional standard Wiener process. Denote $X_t=(w_t^{\top}, \theta_t^{\top})^{\top}\in \mathbb{R}^{2d}$, then again, $X_t$ is a multidimensional OU process with expectation
\begin{equation}
E\left[X(t) | X\left(1\right)\right]=(w_1^{\top} \cos(\ln t)+\theta_1^{\top}\sin(\ln t),w_1^{\top}\sin(\ln t)-\theta_1^{\top}\cos(\ln t))^{\top}
\end{equation}
and variance 
\begin{equation}
\operatorname{Var}\left[X(t) | X\left(1\right)\right]=\int_{1}^{t} e^{a(t)} e^{-a(s)} \Sigma(s) \Sigma(s)^{\top}\left(e^{a(t)} e^{-a(s)}\right)^{\top} d s
\end{equation}
where 
\begin{equation}
A(t)=\begin{pmatrix}
0 & -\frac{1}{t}I_d\\
\frac{1}{t}I_d & 0
\end{pmatrix}
,\Sigma(t)=\begin{pmatrix}
\frac{\sqrt{2}}{\sqrt{m}t}I_d & 0\\
0 & 0
\end{pmatrix}
\end{equation}
and $\tilde{A}(t)$ is a primitive function to $A(t)\operatorname{d}t$. For example, we take:
\begin{equation}
\tilde{A}(t)=\int A(t)=\begin{pmatrix}
0 & -(\ln{t})I_d\\
(\ln{t})I_d & 0
\end{pmatrix}
\end{equation}

So $e^{\tilde{A}(t)}=\begin{pmatrix}
(\cos{\ln{t}})I_d & -(\sin{\ln{t}})I_d\\
(\sin{\ln{t}})I_d & (\cos{\ln{t}})I_d
\end{pmatrix}$ and 
\begin{align}
e^{\tilde{A}(t)}e^{-\tilde{A}(s)}=&\begin{pmatrix}
(\cos{\ln{t}})I_d & -(\sin{\ln{t}})I_d\\
(\sin{\ln{t}})I_d & (\cos{\ln{t}})I_d
\end{pmatrix}\begin{pmatrix}
(\cos{-\ln{s}})I_d & -(\sin{-\ln{s}})I_d\\
(\sin{-\ln{s}})I_d & (\cos{-\ln{s}})I_d
\end{pmatrix}\\
=&\begin{pmatrix}
(\cos{\ln{\frac{t}{s}}})I_d & -(\sin{\ln{\frac{t}{s}}})I_d\\
(\sin{\ln{\frac{t}{s}}})I_d & (\cos{\ln{\frac{t}{s}}})I_d
\end{pmatrix}\\
=&\begin{pmatrix}
(\cos{\ln{\frac{s}{t}}})I_d & (\sin{\ln{\frac{s}{t}}})I_d\\
-(\sin{\ln{\frac{s}{t}}})I_d & (\cos{\ln{\frac{s}{t}}})I_d
\end{pmatrix}\\
=&{e^{\tilde{A}(\frac{s}{t})}}^{\top}
\end{align}
the variance of the $i$-th component of $X(t)$ is
\begin{align}
\operatorname{Var}\left[X(t) | X(0)\right]_{ii}=&\int_{1}^{t} \|\left[e^{\tilde{A}(t)} e^{-\tilde{A}(s)} \Sigma(s)\right]_i\|^2_2  d s\\
=&\int_{1}^{t} \|\left[{e^{\tilde{A}(\frac{s}{t})}}^{\top} \Sigma(s)\right]_i\|^2_2 d s\\
=&\int_{1}^{t} \|\left[{e^{\tilde{A}(\frac{s}{t})}}^{\top}\right]_i \Sigma(s)\|^2_2 d s\\
=&\int_{1}^{t} \frac{1}{m}\|\left[{e^{\tilde{A}(\frac{s}{t})}}^{\top}\right]_i \begin{pmatrix}
\frac{\sqrt{2}}{t}I_d & 0\\
0 & 0
\end{pmatrix}
\|^2_2 d s\\
=&\Theta(\frac{1}{m})
\end{align}
From the definition of $X(t)$ we have $\operatorname{Var}([\theta_t]_i)=\operatorname{Var}[X(t)|X(0)]_{d+i,d+i}=\Theta(\frac{1}{m})$. Hence we complete our proof of the second part.
\end{proof}

\section{Proof of Theorem 3}

\begin{proof}
Since $f$ is L-Lipschitz, we have
\begin{align}
|f(\lambda x_1+(1-\lambda)x_2)-f(x_1)|\leq& L\|x_1-(\lambda x_1+(1-\lambda)x_2)\|\\
=&L(1-\lambda)\|x_1-x_2\| \leq L(1-\lambda)\delta  
\end{align}
\begin{align}
|f(\lambda x_1+(1-\lambda)x_2)-f(x_2)|\leq& L\|x_2-(\lambda x_1+(1-\lambda)x_2)\|\\
=&L\lambda\|x_1-x_2\|\leq L\lambda \delta
\end{align}
So we have
$$f(\lambda x_1+(1-\lambda)x_2)\geq f(x_1)-L(1-\lambda)\delta $$
and
$$f(\lambda x_1+(1-\lambda)x_2)\geq f(x_2)-L\lambda\delta $$
Hence, $f(\lambda x_1+(1-\lambda)x_2)\geq \max\{f(x_1)-L(1-\lambda)\delta,f(x_2)-L\lambda\delta\} $. And the second result follows by simple calculation.

\end{proof}

\section{Sample insufficiency}
\label{ap_si}
We show more examples to demonstrate the problem of sample insufficiency of GANs. It can misguide the discriminator and then generator, causing finally generator to generate anomalous images. The training dataset consists images whose rectangle number is exactly 2. They are 32 by 32 single-channel image. All rectangles are 8 by 8. Anomalous generated images have different number of rectangles. The two training designs are compared. First one is mini-batch gradient descent (MGD), where the insufficient problem is severe and training is unstable, giving rise to anomalous images during whole training. Second is full-batch gradient descent (FGD), where the insufficient problem is negligible. Training for the second approach is stable, with few anomalous images generated. Generated images with certain individual latent codes are plotted. Images are grey single-channel but shown in color for visualization purpose.
\begin{figure}[h]
    \centering
    \includegraphics[width=1.0\linewidth]{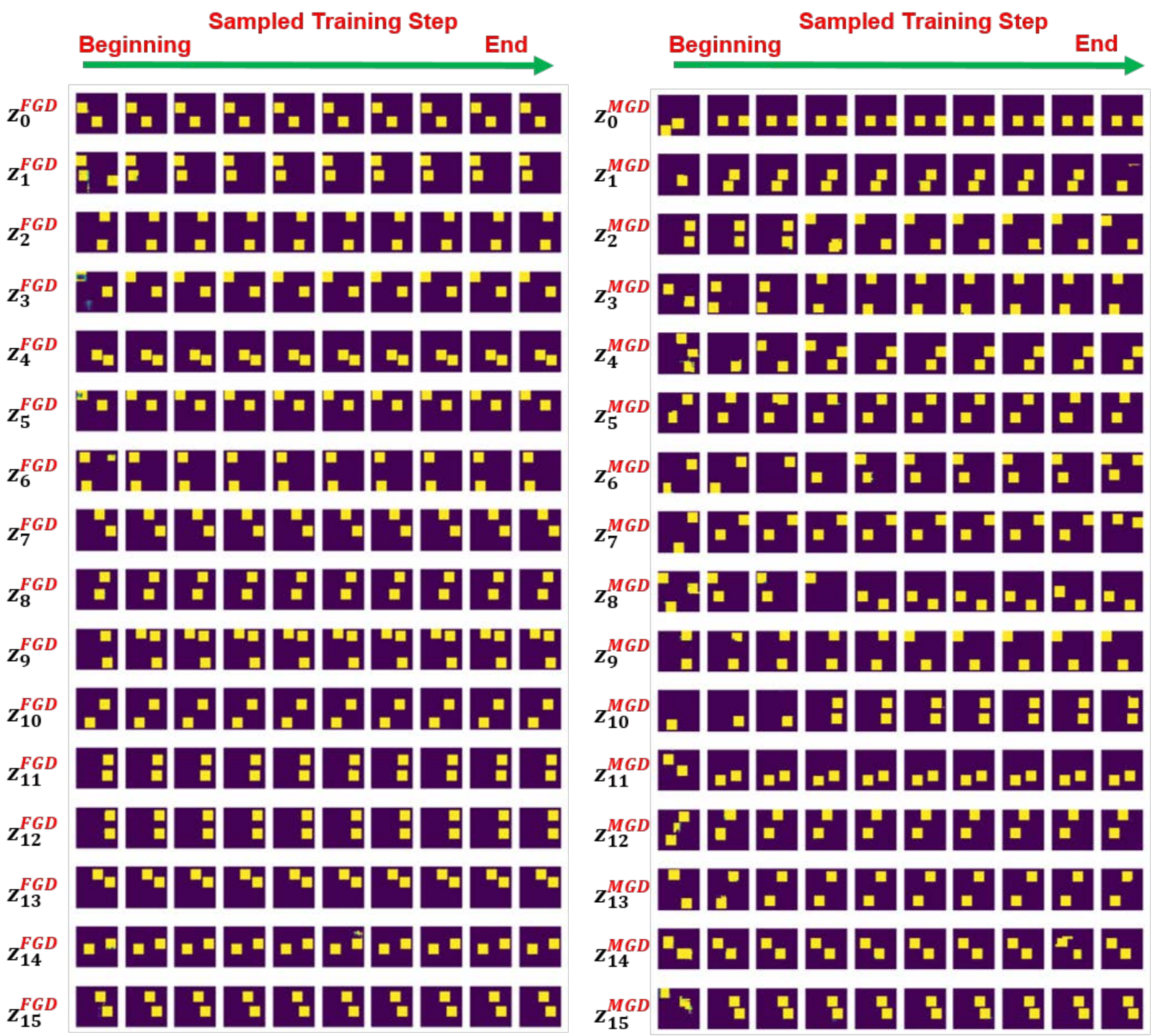}
    \caption{The generated images for certain fixed latent codes. For FGD (left), training is stable and converges smoothly. For MGD, where the insufficient problem is severe, training is unstable and gives rise to anomalous images (rectangle number is not two).}
    \label{ap_asc}
\end{figure}
\section{Avoiding anomalous generalization on geometry data}
\label{ap_aag}
We introduce two problems to explain the anomalous generalization results reported in \cite{zhao2018bias}. Further, we demonstrate the anomalous can be avoided by training modifications. We have three modified training methods extended from the Vanilla training: Sample Correction (SC), Pixel-wise Combination Regularization (PCR) and the SC + PCR. For SC, the negative training samples generated from generator are discarded before fed to discriminator for gradient descent. The samples with true rectangle number is discarded. The selection can be implemented efficiently by counting the number of rectangle using straight forward counting algorithm. For PCR approach, the pixel-wise averages of the training data (pixel-wise logic-and and logic-union) are precomputed. Specifically, the way we generate training geometry data can be utilized for this precomputation. To generate 2-number-rectangle training geometry data, we first randomly generate several 3-number-rectangle images. After that, for each 3-number-rectangle image, we randomly remove one rectangle out of the three. The remove is done twice for each 3-number-rectangle image. By this we get two different 2-number-rectangle images out of one 3-number-rectangle image. Because of this construction method, we could obtain the pixel-wisely average images easily, \emph{i.e.} pixel-wise logic-or or logic-and, which are 3-number-rectangle or 1-number-rectangle respectively. These precomputed additional images are used as additional negative training samples for the PCR approach. All images are 32 by 32 with one channel. All rectangles are 8 by 8.

\begin{figure}[h]
    \centering
    \includegraphics[width=1.0\linewidth]{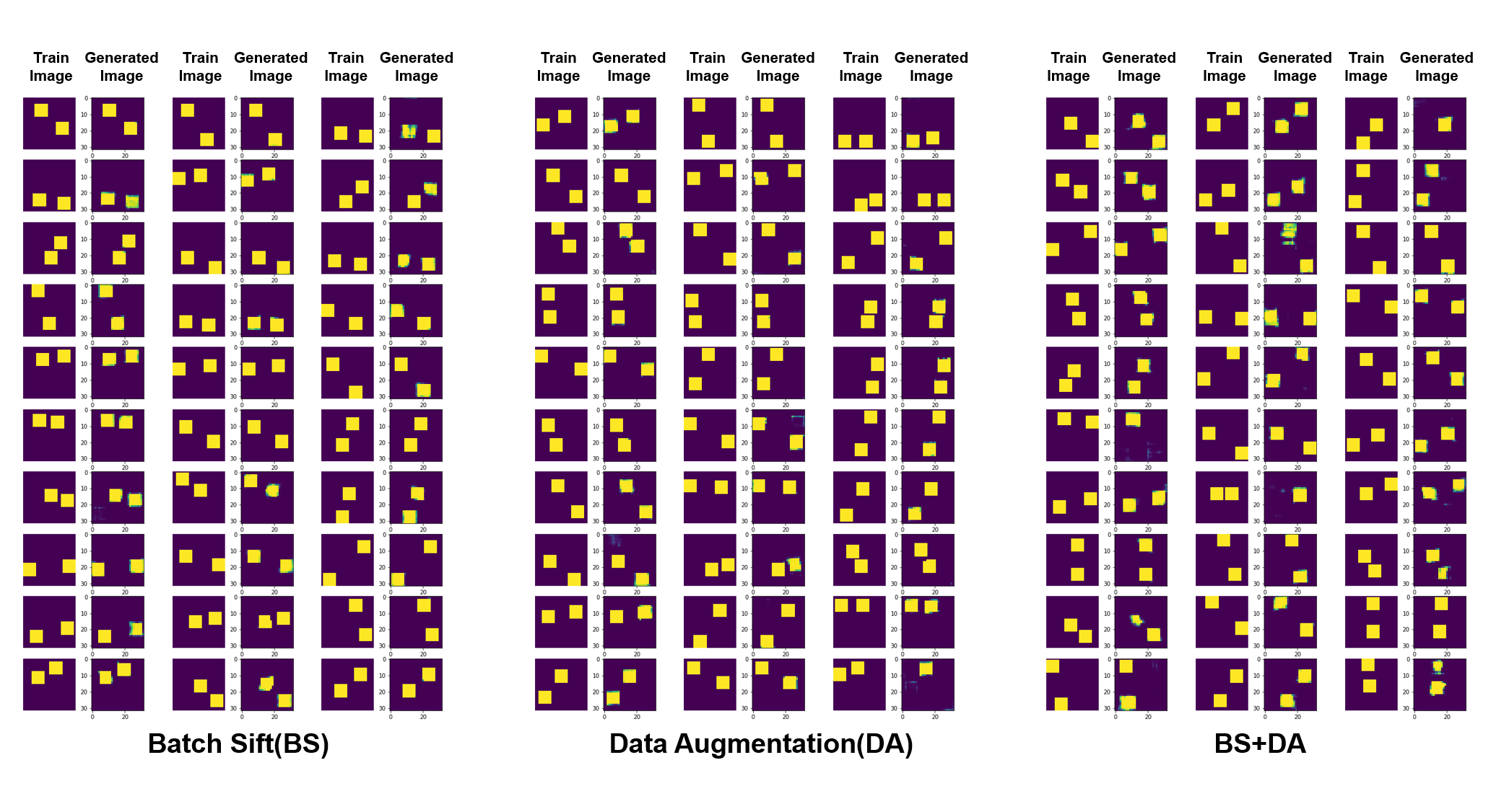}
    \caption{For a random training image, the closest generated image in 256000 generated samples is found. For three training method (SC, PCR and SC+PCR), most training images are represented by the learned distribution of the generator, meaning the high performance is achieved (up to 99\% correct generated images) without mode collapse.}
    \label{ap_nmc}
\end{figure}

In experiments, SC and PCR can both improve the proportion of correct generated images. Combining the SC and PCR, the proportion goes to 100\% quickly. We show there is no mode collapses for these three training modifications: SC, PCR and SC+PCR. We randomly draw 30 training images. For each sample, we find the closest image in 256000 generated samples. Results are shown in Figure \ref{ap_nmc}. For three training designs extended from the Vanilla (SC, PCR and SC+PCR), most training images are represented by the learned distribution of the generator, meaning the high performance is achieved (up to 99\% correct generated images) without mode collapse.

\end{document}